\documentclass{article}

\usepackage{arxiv}

\usepackage[utf8]{inputenc} 
\usepackage[T1]{fontenc}    
\usepackage{hyperref}       
\usepackage{url}            
\usepackage{booktabs}       
\usepackage{amsfonts}       
\usepackage{nicefrac}       
\usepackage{microtype}      
\usepackage{graphicx}
\graphicspath{ {./images/} }

\usepackage[usenames,dvipsnames]{xcolor}
\usepackage{subcaption}
\usepackage[normalem]{ulem}
\usepackage{dblfloatfix}
\usepackage{amsmath,amssymb}

\gdef\orcidlogo{%
\includegraphics{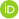}%
}%
\gdef\orcid#1{\href{#1}{\orcidlogo}}%

\title{XAI for Point Cloud Data using Perturbations based on Meaningful Segmentation}

\author{
 \uppercase{Raju Ningappa Mulawade\orcid{https://orcid.org/0000-0002-0180-8517}} \\
  ZFT \& UX-Vis\\
  Hochschule Worms University of Applied Sciences\\
  Erenburgerstr. 19 \\
  Worms, 67549, Germany \\
  \texttt{mulawade@hs-worms.de} \\
   \And
 \uppercase{Christoph Garth\orcid{https://orcid.org/0000-0003-1669-8549}} \\
  Scientific Visualization Lab\\
  RPTU Kaiserslautern-Landau\\
  Gottlieb-Daimler-Str. \\
  Kaiserslautern, 67663, Germany\\
  \texttt{garth@rptu.de} \\
  \And
 \uppercase{Alexander Wiebel\orcid{https://orcid.org/0000-0002-6583-3092}} \\
  ZFT \& UX-Vis\\
  Hochschule Worms University of Applied Sciences\\
  Erenburgerstr. 19 \\
  Worms, 67549, Germany\\
  \texttt{wiebel@hs-worms.de} \\
}

\begin{document}
\date{}
\maketitle
\begin{abstract}
In this work, we propose a novel segmentation-based explainable artificial intelligence (XAI) method for neural networks working on point cloud classification. As one building block of this method, we also propose a novel point-shifting mechanism to introduce perturbations in point cloud data.

In the last decade, Artificial intelligence (AI) has seen an exponential growth. However, due to the "black-box" nature of many of these AI algorithms, it is important to understand their decision-making process when it comes to their application in critical areas. Our work focuses on explaining AI algorithms that classify point cloud data. An important aspect of the methods used for explaining AI algorithms is their ability to produce explanations that are easy for humans to understand. This allows the users to analyze the performance of AI algorithms better and make appropriate decisions based on that analysis. Therefore, in this work, we intend to generate meaningful explanations that can be easily interpreted by humans.
The point cloud data considered in this work represents 3D objects such as cars, guitars, and laptops. We make use of point cloud segmentation models to generate explanations for the working of classification models. The segments are used to introduce perturbations into the input point cloud data and generate saliency maps. The perturbations are introduced using the novel point-shifting mechanism proposed in this work which ensures that the shifted points no longer influence the output of the classification algorithm.

In contrast to any previous methods, the segments used by our method are meaningful, i.e. humans can easily interpret the meaning of these segments. Thus, the benefit of our method over other methods is its ability to produce more meaningful saliency maps. We compare our method with the use of classical clustering algorithms to generate explanations. We also analyze the saliency maps generated for some example inputs using our method to demonstrate the usefulness of our proposed method in generating meaningful explanations.
\end{abstract}


\section{Introduction\label{introduction}}
Explainable artificial intelligence (XAI) has become an important field of research in the last decade. This is mainly due to the exponential growth in AI which is finding use in almost every field of application from agriculture to autonomous vehicles. AI algorithms are now capable of performing various difficult tasks with high accuracy. This has prompted industries belonging to various fields to incorporate AI algorithms and improve the performance of various tasks performed in those industries. However, most AI algorithms performing challenging tasks have complex architecture which makes it highly difficult to understand how the algorithm is making a decision. Therefore, many such AI models are referred to as "black boxes". This is one of the primary concerns related to AI that hinders the use of AI algorithms in high-risk tasks. Thus, as AI algorithms learn to perform more complex tasks, the need to understand their decision-making process becomes more important. Our work contributes to this important field of research.

\begin{figure}[t!]
    \centering
    \includegraphics[width=\linewidth]{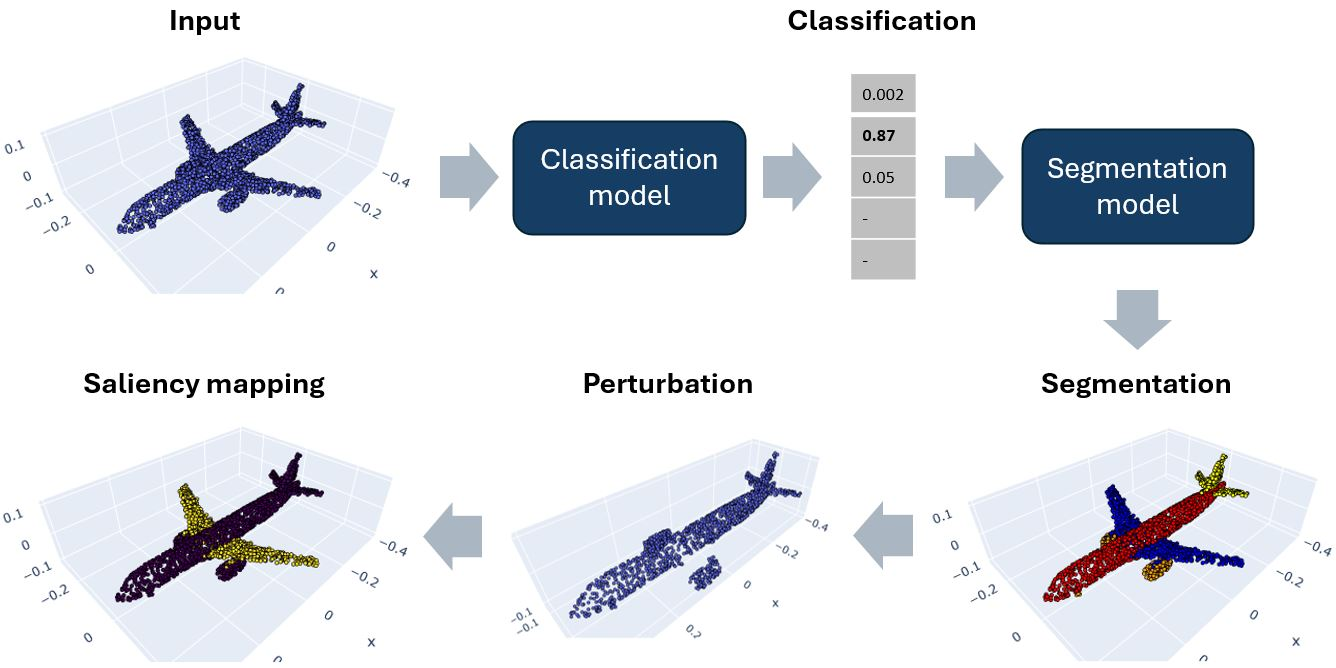}
    \caption{An overview of the XAI pipeline proposed.}
    \label{fig:pipeline}
\end{figure}

AI algorithms work on various types of data such as text, tabular, image, and point clouds. In this work, we propose an explainability method that focuses on explaining the classification models working on point cloud data as we see a growing trend in the use of point cloud data for AI model development in the last decade~\cite{Guo_2021_DL-in-PC}~\cite{Camuffo_2022_An_Updated_Overview}. We also observe a similar trend in XAI research work targeting algorithms working on point cloud data~\cite{Mulawade_2024_XAI_survey}~\cite{5_Zhang_2020_PointHop}~\cite{4_Matrone_2022_BubbleEX}~\cite{13_Zheng_2019_PointCloud}. However, there is still a significant gap between the XAI work developed for data types such as image and text compared to point cloud data. Therefore, through our work, we attempt to reduce this gap by contributing a method based on meaningful segmentation to the point cloud-based XAI field of research. \autoref{fig:pipeline} shows an overview of our proposed method.

The main contributions of our work are:
\begin{itemize}
    \item Segmentation-based XAI for understanding classification networks working on point cloud data. The proposed method is a perturbation-based method and it is model-agnostic.
    \item Proposal of a novel point shifting mechanism for the perturbation of point cloud data.
    \item Two types of introducing perturbations to generate explanations that provide different interesting insights.
    \item Detailed analysis of the proposed method against clustering-based methods to highlight its advantages. The analysis shows how the proposed method generates meaningful explanations.
\end{itemize}

The rest of the paper is organized as follows: Section \ref{related work} gives a detailed overview of the literature that is relevant to our work. Section \ref{segmentation-based xai} describes the proposed XAI method in this work and the perturbation mechanisms utilized for generating saliency maps. It also provides an overview of the data and AI models used in this work.        
Section \ref{discussion} provides a detailed analysis of the proposed method using multiple examples to indicate the usefulness of our method.
Section \ref{conclusion} contains our final remarks regarding the work and the direction in which the future work can progress.

\section{Related Work}\label{related work}
As point cloud data is gaining importance in AI developments, the research related to explaining AI algorithms working on point cloud data has also seen an upward trajectory. Many authors have attempted to provide explanations for these algorithms employing various types of explainability mechanisms. Mulawade et al.~\cite{Mulawade_2024_XAI_survey} have provided a detailed survey of all the XAI literature addressing the issue of explainability for AI models working on point cloud data. Saranti et al.~\cite{Saranti_2024_From_3D} provided a survey focusing specifically on the explainability of graph neural networks (GNNs) working on point cloud data. Among the different types of XAI methodologies proposed in the past, the perturbation-based methods have found greater importance in explaining point cloud-based AI models. This is evident in the list of papers surveyed by the authors in \cite{Mulawade_2024_XAI_survey} with papers proposing perturbation-based XAI methods being the highest in number among the methodologies used. Some of the most prominent perturbation-based methods (considering all types of data such as image, text, and point clouds) are SHapley Additive exPlanations (SHAP)~\cite{Lundberg_2017_SHAP} and Local Interpretable Model-agnostic Explanations (LIME)~\cite{Ribeiro_2016_LIME}. The perturbation-based XAI methods for point cloud data use different types of perturbation to generate explanations for the working of AI models. We describe them below and highlight the need for our work. 

Zheng et al.\cite{13_Zheng_2019_PointCloud} proposed an XAI method that computes saliency maps by introducing perturbation into the input data. The perturbation method used in this work uses the process of moving a specific point to the center of the point clouds to introduce perturbations in the input data. The authors consider the spherical coordinate system to compute the attributions corresponding to the points as they are gradually shifted to the center of the input data.

Taghanaki et al.\cite{34_taghanaki_2020_pointmask} proposed a perturbation-based XAI method for explaining classification networks working on point cloud data. They proposed a method called \emph{PointMask} which learns to mask out points in the input data based on their contribution to the output class score.

Shen et al.\cite{17_Shen_2021_Interpreting} proposed a perturbation-based XAI method for analyzing the classification network working on point cloud data. The authors used Shapley values\cite{shapley_1953_value} to compute saliency maps. The input point cloud is segmented into a fixed number of regions and points belonging to specific regions are moved to the center of the point cloud data to measure the changes in the output target class to generate a saliency map. 

Verbung~\cite{15_Verburg_2022_Exploring} proposed a perturbation-based XAI method for understanding a segmentation model working on point cloud data. The author introduced perturbations into the input data by modifying specific regions (such as the shape of a manually selected part of an object) in the input point cloud data and measuring the effect of this perturbation on the segmentation output. 

Tan et al.\cite{24_Tan_2022_Surrogate} proposed an XAI method that adapts LIME~\cite{Ribeiro_2016_LIME} to explain the decision-making process of classification models working on point clouds. The point cloud data is divided into multiple regions using a clustering method and perturbations are introduced using these clusters to compute saliency maps using the LIME methodology. 

Tan~\cite{25_Tan_2023_Visualizing} proposed another perturbation-based XAI method for point cloud-based AI models that perform a classification task. In this method, the target output class score is maximized by modifying manually selected parts of the input data. The authors made use of autoencoders to encode and generate new input samples.  

Tan \cite{26_Tan_2023_Fractual} also proposed an XAI method for understanding a classification network working on point cloud data that is based on feature ablation. The author proposed removing specific features (identified by the author) from all the data instances in the training dataset and retraining the model on the perturbed data. The change in classification accuracy achieved by the model is then used as an attribution that indicates the importance of the removed features.  

Tan and Kotthaus\cite{31_Tan_2023_Explainability-Aware} used integrated gradients\cite{sundararajan_2017_INTGRAD} to identify critical points in the input point cloud data and use these critical points to perturb the input data.

Miao et al.\cite{23_miao_2023_interpretable} proposed Learnable Randomness Injection (LRI) that provides an explanation for the working of a classification model with point cloud data as its input. The proposed method learns to inject randomness (perturbation) into the input data during the training process taking into consideration the performance of the AI model in classifying the data.

The most recent contribution of Tan\cite{48_tan_2024_flowAM} to the topic proposes an activation-flow-based AM method named Flow AM that makes use of the activation maximization of the output target class and also forces the neurons in the intermediate layers to align their activation values to the values that correspond to actual input instances during this process. 

Atik et al.\cite{42_Atik_2024_xai_for_ML} adapted SHAP for interpretation of the classification model working on photogrammetric point cloud data. The authors mainly focused on the explainability of ensemble classifiers in this work.  

Another adaptation of Shapley values for understanding point cloud-based AI models was proposed by Shen et al.\cite{50_shen_2024_REQN} where the authors divided the input point cloud data uniformly into multiple regions and computed Shapley values. The perturbation method used in this work was the "point shifting" mechanism where the points of some regions are moved to the center of the point cloud data.  

Lavasa et al.\cite{60_Lavasa_2024_metrology} adapted SHAP for analyzing the performance of AI models that predict the accuracy of laser scanning devices.

However, none of the above methods mention or describe using \emph{meaningful} segments to introduce perturbations into the input data unless introduced manually by the developers of the methods. The use of meaningless segments for perturbation leads to the generation of saliency maps with attributions assigned to data points that are difficult to interpret. Furthermore, methods employing the perturbation mechanism by shifting or removing individual points are computationally expensive. In addition to this, we also believe that individual points do not carry important information such as structural information that is crucial information in point clouds. This important information is captured by a set of points. Therefore, the perturbation mechanism should consider using sets of multiple points in the point cloud data to introduce perturbations instead of individual points. Furthermore, the information captured by these sets of points should be meaningful. This leads to the generation of saliency maps that are meaningful, and therefore, easy for humans to interpret. In addition to this, perturbations introduced into the input data should generate an input where specific features have no influence on the output.

In this work, we propose a perturbation-based method that makes use of meaningful segments generated by an algorithm to perturb the input data and compute attributions based on the change in the output value of the target class. We also propose a point-shifting mechanism for introducing perturbations in point cloud data that meets the requirement mentioned above.

\section{Segmentation-based XAI}\label{segmentation-based xai}
The proposed segmentation-based XAI method for point cloud classification models utilizes a segmentation model that generates meaningful segments from the given input point cloud data. An overview of the proposed method is shown in \autoref{fig:pipeline}. It consists of four stages: 
\begin{itemize}
    \item Classification
    \item Segmentation
    \item Perturbation
    \item Saliency mapping
\end{itemize}

In the first stage, the input point cloud data is used as the input for the classification model which predicts the output class of this data. This is the same classification model that we intend to understand in the XAI process. Based on the output class, the corresponding segmentation model is chosen from the list of pre-trained models. In the second stage, the selected segmentation model is used to meaningfully segment the input point cloud data. The resulting segments are used in the third stage to perturb the input data. Using the classification model and the perturbed input data, a saliency map is computed in the final stage of this pipeline.

\subsection{Segmentation}\label{segmentation_mechanism}
In our XAI method, we intend to segment the input point cloud data in a meaningful way. This means that the segments produced by the segmentation mechanism are easy to understand for humans. For example, the segmentation of point cloud data representing a human 3D model into segments that represent the head, hands, legs, and torso. In our work, we use two segmentation mechanisms to divide the input point cloud data into multiple meaningful segments. These are explained below. 

\subsubsection{Segmentation mechanism}\label{use_of_segmentation_models}
\begin{figure}[t]
    \centering
    \begin{subfigure}[b]{0.46\linewidth}
        \centering
        \includegraphics[width=\linewidth]{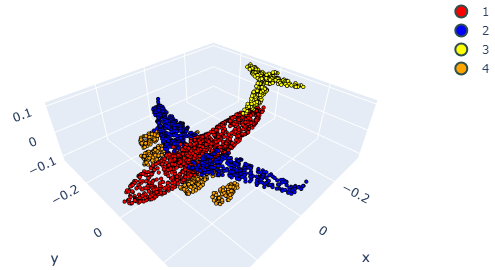}
        \subcaption{\label{fig:airplane_20_seg_GT}Ground truth}
    \end{subfigure}
    \hfill
    \begin{subfigure}[b]{0.46\linewidth}
        \centering
        \includegraphics[width=\linewidth]{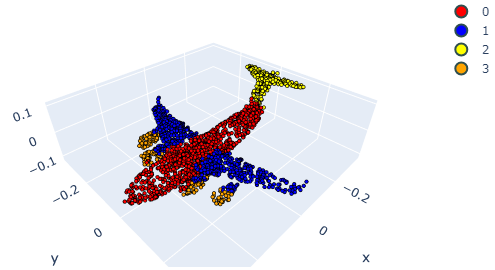}
        \subcaption{\label{fig:airplane_20_seg}Segmentation output}
    \end{subfigure}
    \vfill
    \begin{subfigure}[b]{0.46\linewidth}
        \centering
        \includegraphics[width=\linewidth]{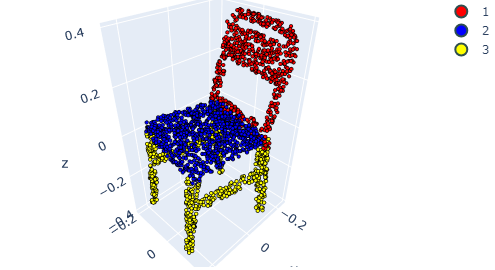}
        \subcaption{\label{fig:chair_01_seg_GT}Ground truth}
    \end{subfigure}
    \hfill
    \begin{subfigure}[b]{0.46\linewidth}
        \centering
        \includegraphics[width=\linewidth]{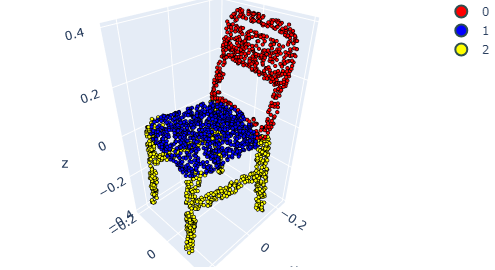}
        \subcaption{\label{fig:chair_01_seg}Segmentation output}
    \end{subfigure}
    \vfill
    \begin{subfigure}[b]{0.46\linewidth}
        \centering
        \includegraphics[width=\linewidth]{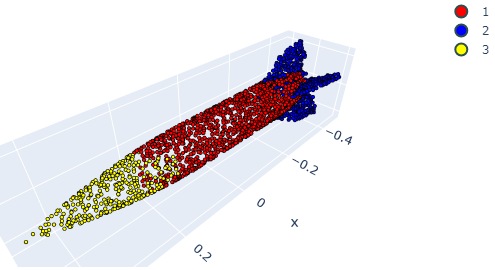}
        \subcaption{\label{fig:rocket_04_seg_GT}Ground truth}
    \end{subfigure}
    \hfill
    \begin{subfigure}[b]{0.46\linewidth}
        \centering
        \includegraphics[width=\linewidth]{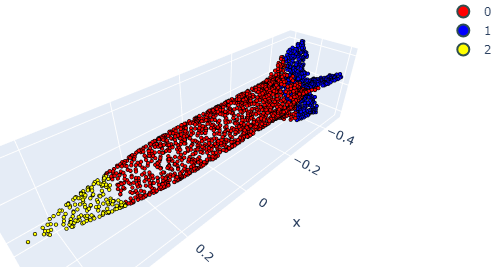}
        \subcaption{\label{fig:rocket_04_seg}Segmentation output}
    \end{subfigure}
    \caption{\label{fig:segmentation_ex}Examples of point clouds segmentation performed by our segmentation models and their corresponding ground truth labels.}
\end{figure}

This mechanism consists of AI models that are trained for part segmentation tasks on the point cloud data. These models use the same input data that is used by the classification model, identify different meaningful segments in the data, and assign them with specific labels. The dataset used in our work consists of point cloud data instances representing 16 types of 3D models. To ensure better performance, we train segmentation models to segment data instances representing a specific 3D model such as airplanes or cars instead of training one single segmentation model to segment point clouds representing every kind of 3D model. Therefore, we have 16 segmentation models with each model catering to segmenting a specific type of point cloud data.
\autoref{fig:segmentation_ex} shows the segmentation of point clouds representing an airplane, a chair, and a rocket by the three corresponding segmentation models.

\subsubsection{Segmentation+Clustering mechanism}\label{sec:segmentation_and_clustering}
The dataset taken into consideration in this work contains point clouds representing various types of 3D models. Some of these models contain features that consist of more than one part, such as the four wheels of cars, the two wheels of motorbikes and the two wings of airplanes. The segmentation algorithms classify these features as one class/group. This leads to perturbations where all parts of these features are shifted (in case of \emph{presence of a feature} mechanism, see~\ref{presence of a feature}) to a chosen point or retained (in case of \emph{absence of a feature} mechanism, see~\ref{absence of a feature}) in the input with remaining segments shifted to the chosen point. This leads to the generation of saliency maps that contain the same saliency attribution value attached to these features belonging to a single class. This can be observed for the wings of airplanes in the saliency maps visualized in \autoref{fig:Airplane_13_sal_map} and \autoref{fig:Airplane_13_sal_map_feat_presence} and their corresponding bar plots \autoref{fig:Airplane_13_sal_map_barplot} and \autoref{fig:Airplane_13_sal_map_feat_presence_bar_plot}. However, the relevance of these multiple features that are grouped into one class is not identical in many cases. Therefore, it can be important to understand the influence these individual features have on the output class score in addition to their influence as a group.

\begin{figure}[t]
    \centering
    \begin{subfigure}[b]{0.48\linewidth}
        \centering
        \includegraphics[width=\linewidth]{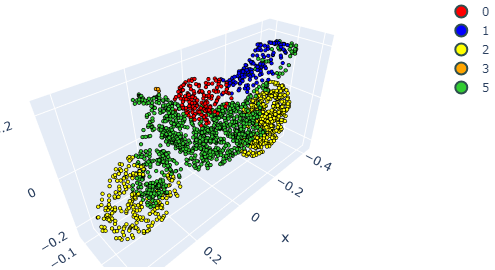}
        \subcaption{\label{fig:seg_clustering_example_a}Segmentation}
    \end{subfigure}
    \hfill
    \begin{subfigure}[b]{0.48\linewidth}
        \centering
        \includegraphics[width=\linewidth]{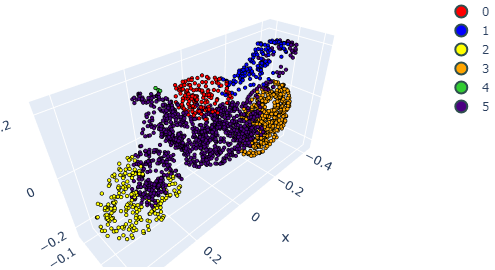}
        \subcaption{\label{fig:seg_clustering_example_b}Segmentation+Clustering}
    \end{subfigure}
    \vfill
    \begin{subfigure}[b]{0.48\linewidth}
        \centering
        \includegraphics[width=\linewidth]{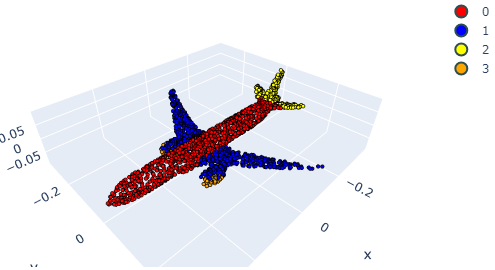}
        \subcaption{\label{fig:seg_clustering_example_c}Segmentation}
    \end{subfigure}
    \hfill
    \begin{subfigure}[b]{0.48\linewidth}
        \centering
        \includegraphics[width=\linewidth]{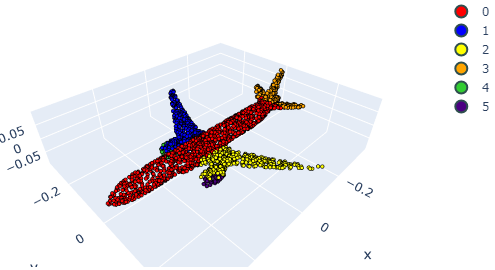}
        \subcaption{\label{fig:seg_clustering_example_d}Segmentation+Clustering}
    \end{subfigure}
    \caption{\label{fig:seg_clustering_example}Clustering of the segments produced by the segmentation models to obtain finer segments.}
\end{figure}

To generate saliency maps for individual features, we made use of clustering algorithms for the segmentation-based mechanism to further segment the input point cloud data. The segmentation model's output is used by the clustering algorithms to further cluster the data. We use two clustering algorithms: 1) DBSCAN~\cite{ester_1996_DBSCAN} for determining the number of clusters in a given segment of the segmentation output. 2)~KMeans~\cite{macqueen_1967_KMeans} clustering algorithm to cluster the given segment based on the number of clusters determined by DBSCAN. We use this combination because the DBSCAN algorithm tends to classify some of the outlying points of the segment as outliers and assigns a separate value to them. This leads to undesired clusters. To avoid this, we combine it with KMeans which takes the number of clusters (without taking outliers' class into account) as input and produces the desired number of clusters. \autoref{fig:seg_clustering_example} shows examples of the segmentation of point clouds representing a motorbike and an airplane using the segmentation and segmentation+clustering methods. \autoref{fig:seg_clustering_example_a} is the output of the segmentation model that identifies the wheels of the motorbike as one segment. Similarly, the wings and engines are labeled as a single segment each, as shown in \autoref{fig:seg_clustering_example_c}. The segmentation+clustering algorithm clusters the wheels of the motorbike to produce front and rear wheels as shown in \autoref{fig:seg_clustering_example_b}. The method also clusters the wings of the airplane into two separate clusters. A similar result is observed with respect to the engines of the airplane as shown in \autoref{fig:seg_clustering_example_d}. 

\subsection{Perturbation mechanism}\label{perturbation_mechanism}
As mentioned in \autoref{introduction}, the proposed XAI method in this work is a perturbation-based method. We use two types of perturbation introducing mechanisms to generate saliency attributions providing interesting insights into the working of the classification model. We explain these mechanisms and the rationale behind them below. 
\subsubsection{Absence of a Feature}\label{absence of a feature}
\begin{figure}[t]
    \centering
    \begin{subfigure}[b]{0.48\linewidth}
        \centering
        \includegraphics[width=\linewidth]{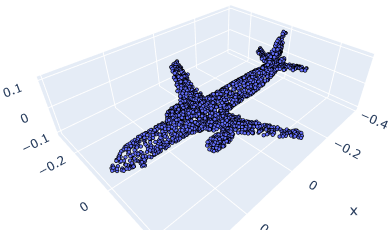}
        \subcaption{\label{fig:Airplane_13}Input}
    \end{subfigure}
    \hfill
    \begin{subfigure}[b]{0.48\linewidth}
        \centering
        \includegraphics[width=\linewidth]{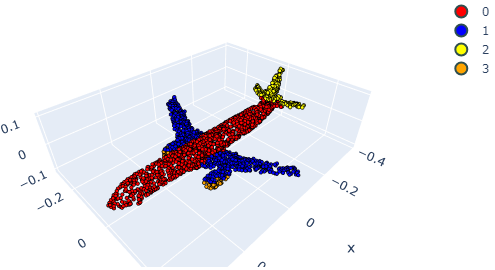}
        \subcaption{\label{fig:Airplane_13_seg}Segmentation output}
    \end{subfigure}
    \vfill
    \begin{subfigure}[b]{0.48\linewidth}
        \centering
        \includegraphics[width=\linewidth]{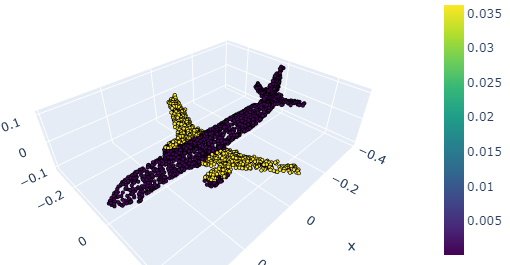}
        \subcaption{\label{fig:Airplane_13_sal_map}Saliency map}
    \end{subfigure}
    \hfill
    \begin{subfigure}[b]{0.45\linewidth}
        \centering
        \includegraphics[width=\linewidth]{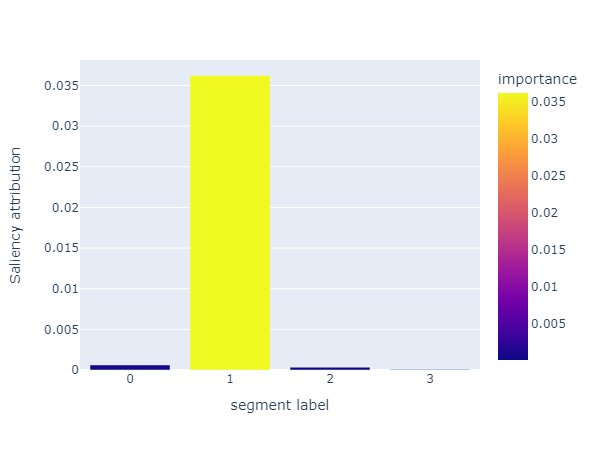}
        \subcaption{\label{fig:Airplane_13_sal_map_barplot}Saliency attributions}
    \end{subfigure}
    \caption{\label{fig:absence_of_feature}Saliency map produced by our method for the given input point cloud data representing a plane. Note: Refer to the 'segmentation output' figure (b) for corresponding parts represented along the x-axis in the bar plot (d).}
\end{figure}

The perturbation mechanism used in this work introduces perturbation by \emph{removing} a specific segment from the input data. Removing here refers to shifting all the points belonging to this specific segment to a chosen point in the input data. A segment can be an individual feature (e.g. bonnet in a car) or a collection of similar features (e.g. wheels of a car).
The perturbed data instance is then used as input for the classification model to compute saliency attributions. The saliency attributions are computed as follows:
\begin{equation}\label{sal_attr}
    S_{AF}(x) = {\lvert P(a)-P(a')\rvert}     
\end{equation}
where $S_{AF}(x)$ is the saliency attribution corresponding to the segment $x$ that is used for perturbation, $a$ is the actual input, $a'$ is the perturbed input, and $P(s)$ refers to the output class score by the classification model for a given input $s$.
\autoref{fig:absence_of_feature} shows an example of this saliency mapping method for input point cloud data representing an airplane.

\subsubsection{Presence of a Feature}\label{presence of a feature}
In addition to the above-mentioned method, we propose a variation of it where we analyze the impact of individual features (or segments) on the output data. Here, we introduce perturbation into the input data by \emph{retaining} a specific segment and moving all the points belonging to other segments to the center of the point cloud data. Mathematically, it can be expressed as follows:
\begin{equation}\label{sal_attr_2}
    S_{PF}(x) = {-\lvert(P(a)-P(a"))\rvert}   
\end{equation}
where $S_{PF}(x)$ is the saliency attribution corresponding to the segment $x$, $a$ is the actual input, $a"$ is the perturbed input where the points not belonging to the segment $x$ are moved to the chosen point, and $P(s)$ refers to the output class score by the classification model for a given input $s$. The minus sign~($-$) is used for the visualization purpose. It allows the segment having the highest influence on the output value to have the highest attribution while the lowest influential segment has the lowest value. 

\begin{figure}[tb]
    \centering
    \begin{subfigure}[b]{0.32\linewidth}
        \centering
        \includegraphics[width=\linewidth]{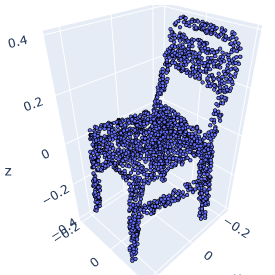}
        \subcaption{\label{fig:chair_01}Input data}
    \end{subfigure}
    \hfill
    \begin{subfigure}[b]{0.32\linewidth}
        \centering
        \includegraphics[width=\linewidth]{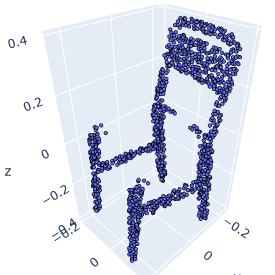}
        \subcaption{\label{fig:chair_01_perturbed_data}Seat part moved}
    \end{subfigure}
    \hfill
    \begin{subfigure}[b]{0.32\linewidth}
        \centering
        \includegraphics[width=\linewidth]{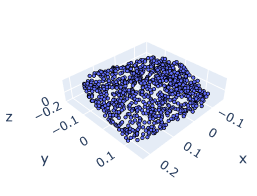}
        \subcaption{\label{fig:chair_01_perturbed_data_2}Seat part}
    \end{subfigure}
    \caption{\label{fig:feat_presence_example} Structural information carried by a point cloud data (left figure) and its perturbed data where all the points belonging to one segment (right figure) are moved to the center of the data (center figure).}
\end{figure}

\begin{figure}[bp]
    \centering
    \begin{subfigure}[b]{0.48\linewidth}
        \centering
        \includegraphics[width=\linewidth]{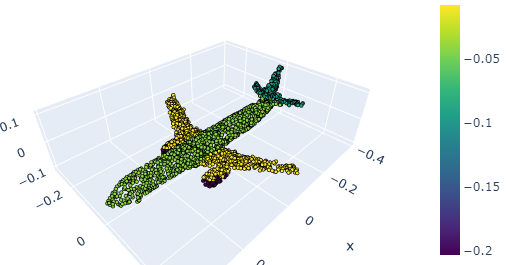}
        \subcaption{\label{fig:Airplane_13_sal_map_feat_presence}Saliency map}
    \end{subfigure}
    \hfill
    \begin{subfigure}[b]{0.45\linewidth}
        \centering
        \includegraphics[width=\linewidth]{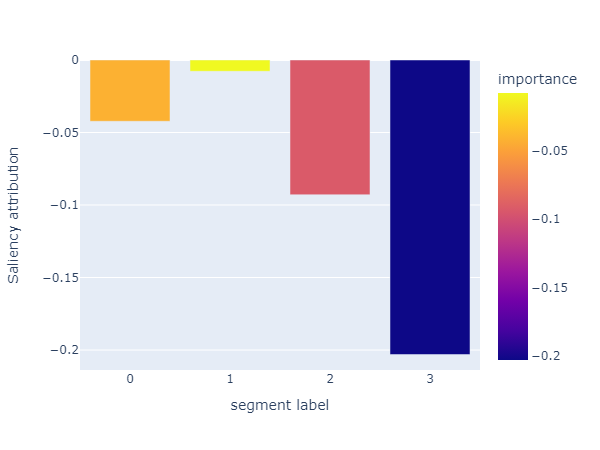}
        \subcaption{\label{fig:Airplane_13_sal_map_feat_presence_bar_plot}Saliency attributions}
    \end{subfigure}
    \caption{\label{fig:presence_of_feature}Saliency map produced by using the "presence of feature" method for the input used in \autoref{fig:absence_of_feature}. Note: Refer to the 'segmentation output' figure (b) in \autoref{fig:absence_of_feature} for corresponding parts represented along the x-axis in the bar plot.}
\end{figure}

The saliency attributions generated by this method can be interpreted as a measure of the influence an individual segment has on the output prediction value when it is the only segment present in the input. This informs us about how good a specific segment is in carrying crucial information on its own. This interpretation is slightly different from the previous one (described in \ref{absence of a feature}) as it does not provide the model with any other information that is carried by other segments or the information that is generated when we combine two or more segments as shown in \autoref{fig:feat_presence_example} where the perturbed data (\autoref{fig:chair_01_perturbed_data}) manages to capture the structure of chair even after the points belonging to one segment (\autoref{fig:chair_01_perturbed_data_2}) are moved to the center of the data. Therefore, we decided to look into how much information a single segment carries that is independent of all the other segments. \autoref{fig:presence_of_feature} shows the saliency attributions computed using this perturbation mechanism for the same input point cloud as data considered in \autoref{fig:absence_of_feature}.

\subsection{Data and Model}\label{data and model}
The dataset used in this work contains the part segmentation of a subset of ShapeNetCore\cite{Chang_2015_shapenet} models based on the work of Yi et al.\cite{yi_2016_scalable}. The dataset consists of $\sim$16000 models from 16 shape categories and the number of data instances in each category varies from 55 to 5266. The number of parts for each model in each category also varies from two to six as each category consists of different types of 3D models representing a specific object such as an airplane. We use this dataset for both classification and segmentation tasks. 

We use classification and segmentation networks based on the PointNet\cite{7_Qi_2017_PointNet} architecture. The classification model is trained on the above-mentioned dataset containing point clouds of 16 different categories. We trained two classification models, one with the default orientation of the point cloud objects in the dataset and the other with the augmented dataset where we modify the orientation of the point cloud objects. We trained the former classification model (that uses the data with default orientation) for 10 epochs with the stochastic gradient descent (SGD) optimizer at 0.001 learning rate and the latter (with augmented data) for 100 epochs (because of the increased complication in the input data due to the augmentation) while keeping the remaining hyperparameters unchanged.

To obtain better segmentation results, we trained individual models to segment particular point cloud data types. Our dataset consists of point clouds representing 16 types of 3D objects such as airplanes, tables, and cars. Thus, we trained 16 segmentation models with each model focusing on segmenting the point cloud data representing a specific 3D object. We also augmented the training data for these segmentation models by modifying the orientation of the data instances. This is similar to the augmentation performed on the training data for classification models. We trained these models with SGD optimizer at 0.001 learning rate with the number of epochs varying from 20 to 150 depending on the size of the subsets representing the 3D objects. \autoref{fig:segmentation_ex} shows some examples that demonstrate the performance of our trained segmentation models on their corresponding input data.

The output of these segmentation models will be used to introduce perturbations into input data instances as described above. We use the point-shifting mechanism to introduce perturbations as it allows the input point cloud instance to retain its number of points thereby avoiding complications in the saliency method. The point-shifting mechanism is described in the following subsection.

\subsection{Point Shifting Mechanism}\label{point_shifting_mechanism}
\begin{figure}[btp]
    \centering
    \begin{subfigure}[b]{0.325\linewidth}
        \centering
        \includegraphics[width=\linewidth]{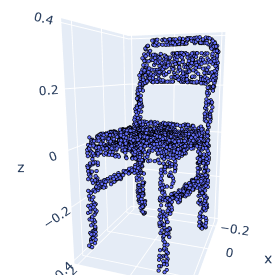}
        \subcaption{\label{fig:chair_01_ps}Input point cloud}
    \end{subfigure}
    \hfill
    \begin{subfigure}[b]{0.325\linewidth}
        \centering
        \includegraphics[width=\linewidth]{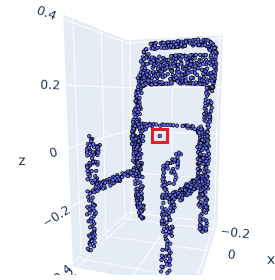}
        \subcaption{\label{fig:chair_01_seat_absence_marked}Perturbed input}
    \end{subfigure}
    \hfill
    \begin{subfigure}[b]{0.325\linewidth}
        \centering
        \includegraphics[width=\linewidth]{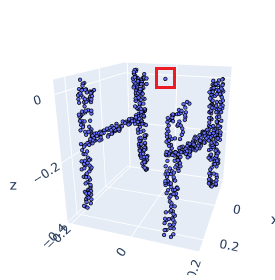}
        \subcaption{\label{fig:chair_01_legs_presence_marked}Perturbed input}
    \end{subfigure}
    \caption{\label{fig:point_shifting_need}Shifting points to the centroid of the input point cloud data. Points representing the seat (in (b)) and seat \& backrest (in (c)) are shifted to the center of the input data (marked with red rectangle).}
\end{figure}

To introduce perturbations into the input data, as mentioned in \autoref{data and model}, we use the point-shifting mechanism. Zheng et al.\cite{13_Zheng_2019_PointCloud} proposed the idea of shifting the points to the center of the input data instead of dropping them from the input. This is based on the intuition that all the outward points in the point cloud determine the output class score of the classification model as they encode shape information while the points closer to the center of the point cloud have almost no influence. However, this process of shifting the points to the center of the point clouds does not fit well with our work. For example, \autoref{fig:point_shifting_need} shows an example of the perturbation of the input data by moving the points belonging to two segments (seat and backrest of the chair) to the center (marked by a red rectangle) of the input data. Thus, the center of the input data now contains a large number of input points and is not actually a part of the retained structure. Therefore, it can act as an additional feature in the input data which is undesirable as we expect the shifted points to have no influence on the decision-making process.  

To address this issue, we need to determine a point in the input space where the points belonging to the specific segment can be shifted, and the shifted points do not influence the output class score. This is possible when the shifted points do not add any structural information to the data. Shifting the points to the center of the retained structure does not always fulfill this requirement. This is evident in \autoref{fig:chair_01_legs_presence_marked} where the center of the retained structure (the legs of the chair) lies in between the leg structures and thus acts as an additional structure in the perturbed data. 

One feasible solution is when the selected point for shifting the points is itself a part of the retained structure in the perturbed input data. This will allow the shifted points to be a part of the perturbed data and provide no additional structural information for the classification model. Since we have multiple points in the retained structure in the input data, we choose a random point from it for shifting the points to. We observe that the saliency attributions corresponding to the features do not vary when selecting random points for perturbation. We discuss this mechanism with some examples in \autoref{discussion}.

\section{Results and Discussion}\label{discussion}
In this section, we evaluate our method using various examples and criteria to highlight the usefulness of the mechanisms that are part of our proposed method.


\subsection{Clustering-based method}
In this subsection, we analyze the use of classical clustering algorithms for segmenting point cloud data, indicate the issues associated with their use, and describe how our method overcomes these issues. 
For the analysis, we used clustering algorithms such as the $k$-means algorithm to generate clusters in the input point cloud data and use these clusters to perturb the same input data to compute saliency attributions. 
\begin{figure}[tbp]
    \centering
    \begin{subfigure}[b]{0.48\linewidth}
        \centering
        \includegraphics[width=\linewidth]{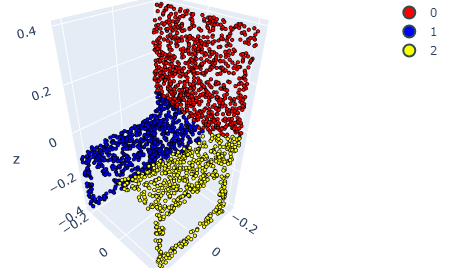}
        \subcaption{\label{fig:chair_2_clusters_3}Segments for c=3}
    \end{subfigure}
    \hfill
    \begin{subfigure}[b]{0.48\linewidth}
        \centering
        \includegraphics[width=\linewidth]{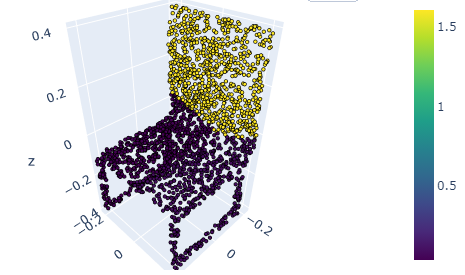}
        \subcaption{\label{fig:chair_2_clusters_3_sal_map}Saliency map for c=3}
    \end{subfigure}
    \vfill
    \begin{subfigure}[b]{0.48\linewidth}
        \centering
        \includegraphics[width=\linewidth]{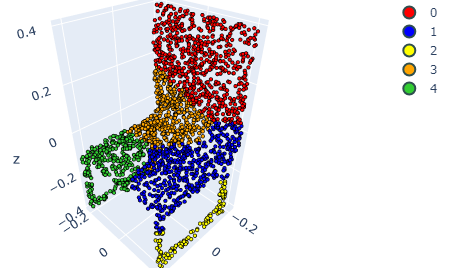}
        \subcaption{\label{fig:chair_2_clusters_5}Segments for c=5}
    \end{subfigure}
    \hfill
    \begin{subfigure}[b]{0.48\linewidth}
        \centering
        \includegraphics[width=\linewidth]{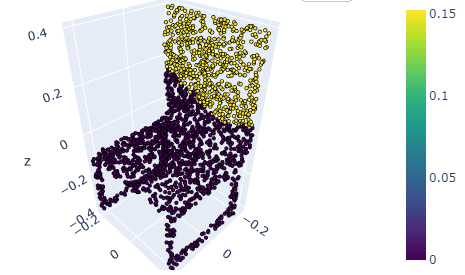}
        \subcaption{\label{fig:chair_2_clusters_5_sal_map}Saliency map for c=5}
    \end{subfigure}
    \vfill
    \begin{subfigure}[b]{0.48\linewidth}
        \centering
        \includegraphics[width=\linewidth]{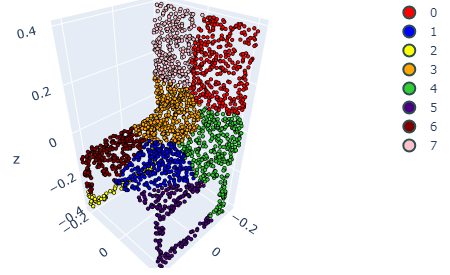}
        \subcaption{\label{fig:chair_2_clusters_8}Segments for c=8}
    \end{subfigure}
    \hfill
    \begin{subfigure}[b]{0.45\linewidth}
        \centering
        \includegraphics[width=\linewidth]{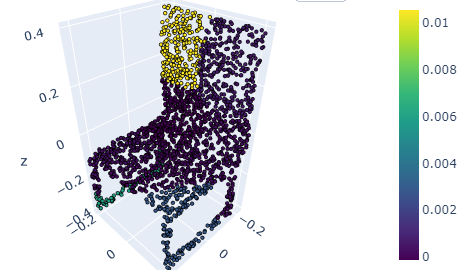}
        \subcaption{\label{fig:chair_2_clusters_8_sal_map}Saliency map for c=8}
    \end{subfigure}
    \vfill
    \begin{subfigure}[b]{0.48\linewidth}
        \centering
        \includegraphics[width=\linewidth]{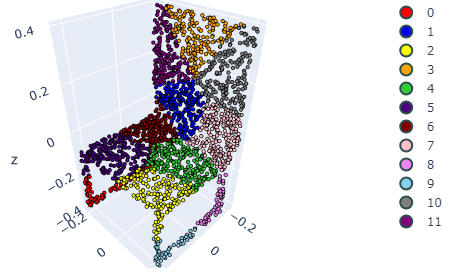}
        \subcaption{\label{fig:chair_2_clusters_12}Segments for c=12}
    \end{subfigure}
    \hfill
    \begin{subfigure}[b]{0.48\linewidth}
        \centering
        \includegraphics[width=\linewidth]{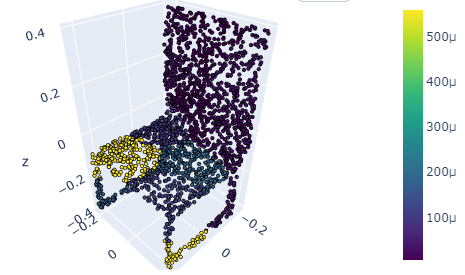}
        \subcaption{\label{fig:chair_2_clusters_12_sal_map}Saliency map for c=12}
    \end{subfigure}
    \caption{\label{fig:clustering_example}Saliency maps produced by the \emph{absence of feature} mechanism for clusters produced by KMeans clustering method. $c$ represents the number of clusters.}
\end{figure}
\autoref{fig:clustering_example} shows examples of using the KMeans clustering method with varying numbers of clusters, $c$, for generating segments for computing the saliency maps. We used the \emph{absence of feature} mechanism to introduce perturbations and compute saliency attributions. We observed that the saliency maps differ as we vary the number of clusters. The most important part for $c=3$ is the top part of the chair, and as we increase $c$ to 12, the most important part shifts to the front left corner of the seat and the front bottom part of the right leg of the chair. We observed similar behavior with other clustering methods such as \textit{spectral} and \textit{agglomerative clustering}. This makes the use of clustering methods for XAI methods tailored for point cloud data unreliable. Our proposed method addresses this issue by using segmentation models that are trained to segment a given point cloud data into a specific number of segments. Another major advantage of using segmentation models over classical clustering algorithms is their ability to learn and adapt to new types of data instances. In other words, we can improve the performance of segmentation models by training it on more data whereas the classical clustering algorithms do not offer this flexibility. 

\subsection{Use of Random point}\label{center_vs_random_point}
As mentioned in \autoref{point_shifting_mechanism}, it is important to find a point in the input space where the points belonging to selected segments can be shifted, and these shifted points do not add any structural information to the perturbed data. In this section, we use an example to discuss and understand how our proposed method of selecting a random point in the retained structure yields better results than other methods such as moving the points to the origin or to the center of the point cloud data.

\begin{figure}[tp]
    \centering
    \begin{subfigure}[b]{0.48\linewidth}
        \centering
        \includegraphics[width=\linewidth]{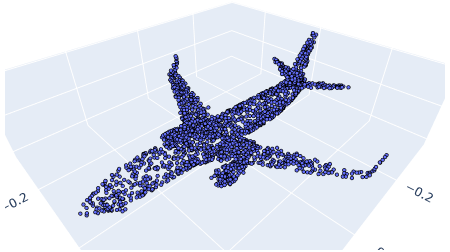}
        \subcaption{\label{fig:seg_and_clustering_analysis_a}Input}
    \end{subfigure}
    \hfill
    \begin{subfigure}[b]{0.48\linewidth}
        \centering
        \includegraphics[width=\linewidth]{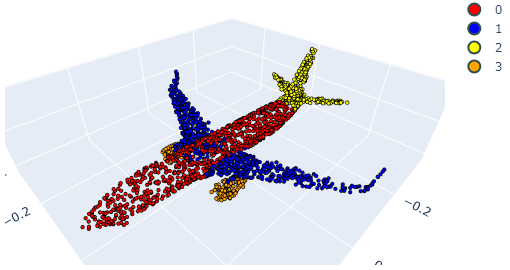}
        \subcaption{\label{fig:seg_and_clustering_analysis_b}Segmentation output}
    \end{subfigure}
    \vfill
    \begin{subfigure}[b]{0.48\linewidth}
        \centering
        \includegraphics[width=\linewidth]{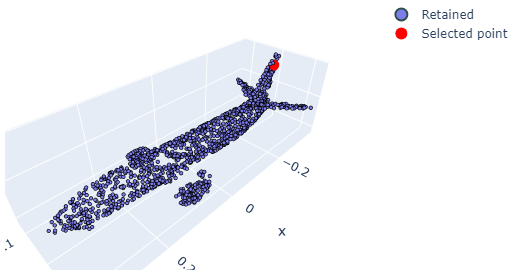}
        \subcaption{\label{fig:seg_and_clustering_analysis_c}Perturbation using random point 1}
    \end{subfigure}
    \hfill
    \begin{subfigure}[b]{0.48\linewidth}
        \centering
        \includegraphics[width=\linewidth]{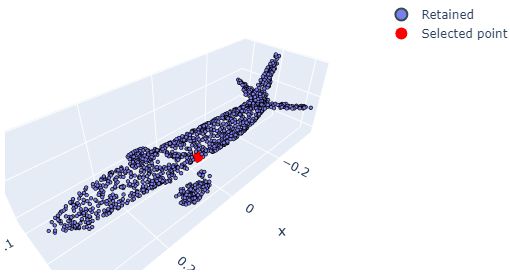}
        \subcaption{\label{fig:seg_and_clustering_analysis_d}Perturbation using random point 2}
    \end{subfigure}
    \caption{\label{fig:seg_and_clustering_analysis}Perturbation of the input data using random points (marked in (c) and (d)) selected from the retained structure.}
\end{figure}

\autoref{fig:seg_and_clustering_analysis} shows examples of input data perturbation for an input data representing an airplane. \autoref{fig:seg_and_clustering_analysis_b} is the segmentation output obtained from a segmentation model which is used to introduce perturbations into the input data shown in \autoref{fig:seg_and_clustering_analysis_a}. For this example, we selected the segment representing the wings to introduce perturbations. We selected a random point in the retained structure (structure without wings) and shifted the points belonging to the segment representing wings. Two examples are shown in \autoref{fig:seg_and_clustering_analysis} with one random point selected in the tail region of the airplane (see \autoref{fig:seg_and_clustering_analysis_c}) while the other random point is selected in the central part of the fuselage (see \autoref{fig:seg_and_clustering_analysis_d}). We shift the points representing the wings to these random points and use these two perturbed data instances to analyze the effect of the perturbations. We use them as input for the classification model and study the change in the output values. We observed that the output values (all 16 values in the output vector) did not change. In other words, the choice of point in the retained structure had no influence on the output values. We observed a similar pattern when we chose different points in the retained structure to shift the points. This observation strengthened our intuition that when the shifted points are a part of the retained structure (irrespective of the point selected in the retained structure for the shifting process), they do not provide any additional structural information for the classification model. Therefore, we use the random point selection mechanism for our point-shifting process.

\subsection{Effects of Different Feature Instances: Wings vs. Fuselage}\label{wings_vs_fuselage}
The datasets used for training classification models usually contain a large number of samples representing different classes. In addition to the differences between the samples representing each class, samples representing a specific class also vary slightly with respect to the information they carry. One such example from our dataset is the use of point clouds representing airplanes with varying numbers of engines on the wings. In this section, we analyze the saliency maps generated by our method to understand how different numbers of engines on the wings affect the output class score. These saliency maps are generated using the classification model that was trained on the dataset containing point clouds in the default orientation.

\begin{figure}[b]
    \centering
    \begin{subfigure}[b]{0.245\linewidth}
        \centering
        \includegraphics[width=\linewidth]{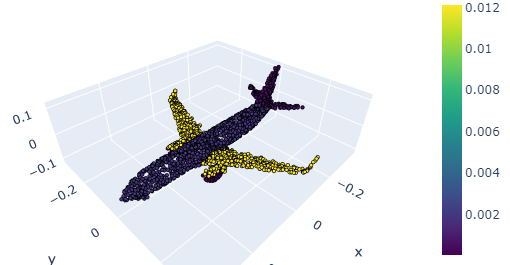}
        \subcaption{\label{fig:wings_vs_body_example_a}}
    \end{subfigure}
    \hfill
    \begin{subfigure}[b]{0.245\linewidth}
        \centering
        \includegraphics[width=\linewidth]{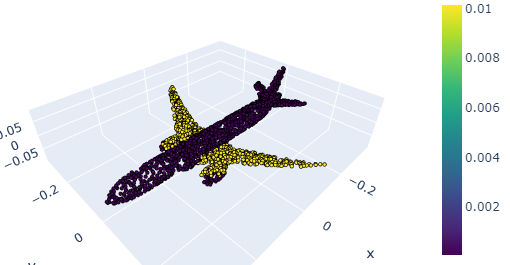}
        \subcaption{\label{fig:wings_vs_body_example_b}}
    \end{subfigure}
    \hfill
    \begin{subfigure}[b]{0.245\linewidth}
        \centering
        \includegraphics[width=\linewidth]{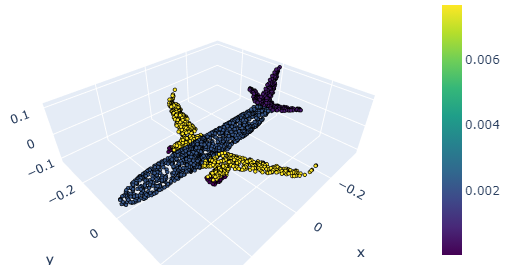}
        \subcaption{\label{fig:wings_vs_body_example_c}}
    \end{subfigure}
    \hfill
    \begin{subfigure}[b]{0.245\linewidth}
        \centering
        \includegraphics[width=\linewidth]{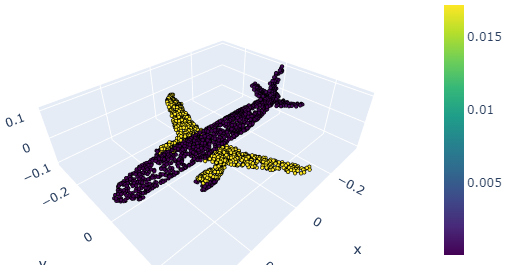}
        \subcaption{\label{fig:wings_vs_body_example_d}}
    \end{subfigure}
    \vfill
    \begin{subfigure}[b]{0.245\linewidth}
        \centering
        \includegraphics[width=\linewidth]{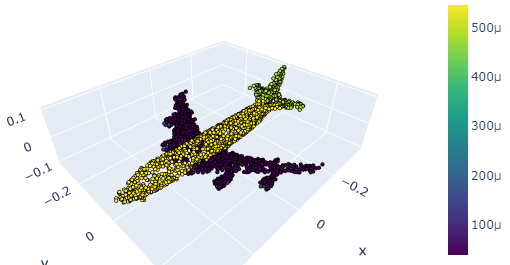}
        \subcaption{\label{fig:wings_vs_body_example_e}}
    \end{subfigure}
    \hfill
    \begin{subfigure}[b]{0.245\linewidth}
        \centering
        \includegraphics[width=\linewidth]{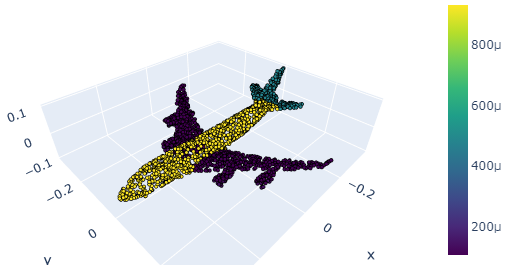}
        \subcaption{\label{fig:wings_vs_body_example_f}}
    \end{subfigure}
    \hfill
    \begin{subfigure}[b]{0.245\linewidth}
        \centering
        \includegraphics[width=\linewidth]{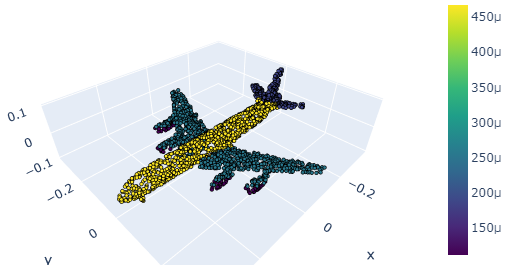}
        \subcaption{\label{fig:wings_vs_body_example_g}}
    \end{subfigure}
    \hfill
    \begin{subfigure}[b]{0.245\linewidth}
        \centering
        \includegraphics[width=\linewidth]{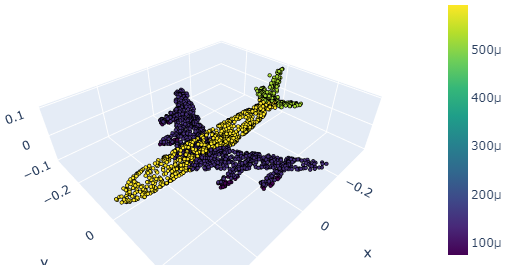}
        \subcaption{\label{fig:wings_vs_body_example_h}}
    \end{subfigure}
    \caption{\label{fig:wings_vs_body_example} Saliency maps of point clouds representing airplanes using the \emph{absence of feature} mechanism.}
\end{figure}

In \autoref{fig:wings_vs_body_example}, we have saliency maps for eight input point cloud data instances representing airplanes. The saliency maps are generated using the \emph{absence of feature} mechanism and the segments are generated using the segmentation models (no clustering). We observe that the saliency maps indicate that the wings are the strongest features for examples in the top row (\autoref{fig:wings_vs_body_example_a}, \autoref{fig:wings_vs_body_example_b}, \autoref{fig:wings_vs_body_example_c}, \autoref{fig:wings_vs_body_example_d}), while the fuselage is the most important feature for the classification model in the bottom row (\autoref{fig:wings_vs_body_example_e}, \autoref{fig:wings_vs_body_example_f}, \autoref{fig:wings_vs_body_example_g}, \autoref{fig:wings_vs_body_example_h}). A more detailed examination of these samples shows that the major difference between the examples on the top row and bottom row is the number of engines. The examples in the top row have two engines whereas the examples in the bottom row have four engines. The shifting of wings to a selected point in the top four examples leads to the two engines adding minute information to the retained structure as these engines are located very close to the fuselage. However, in the remaining examples, the shifting of wings to the selected point leads to a structure with two engines on each side of the fuselage. This captures more information with respect to the overall structure of the airplane and thus leads to a lower influence of wings on the output class value. This is also evident in the color scales of the figures that use saliency values to indicate how influential the features of airplanes are. We observe that the magnitude with which wings affect the output class score is significantly higher compared to other features in the top row's examples. However, the presence of two additional engines in the bottom row's examples brings this magnitude down significantly and leads to the fuselage's influence becoming the biggest among all the features.

This example gives us an important insight into how the classification model learns to identify and take into consideration different structural information captured by point clouds representing a specific object such as an airplane and make decisions based on this information.

\subsection{Segmentation+Clustering: Use case}\label{seg_cluster_usecase}
As described in \autoref{sec:segmentation_and_clustering}, the Segmentation+Clustering mechanism allows the users to analyze a classification model by taking individual features into account instead of using a set of similar features as one segment. This is useful in understanding the contribution of individual features because a set of features can provide more structural information for the classification model compared to individual features. For example, a set of wheels in a motorbike or car captures more information than a single wheel. We analyze an example to see if this is reflected in the saliency maps produced by our methods.

\begin{figure}[btp]
    \centering
    \begin{subfigure}[b]{0.48\linewidth}
        \centering
        \includegraphics[width=\linewidth]{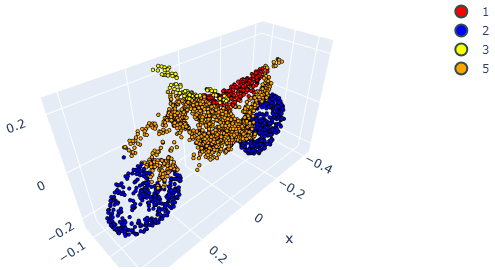}
        \subcaption{\label{fig:seg_cluster_usecase_example_a} Segmentation}
    \end{subfigure}
    \hfill
    \begin{subfigure}[b]{0.48\linewidth}
        \centering
        \includegraphics[width=\linewidth]{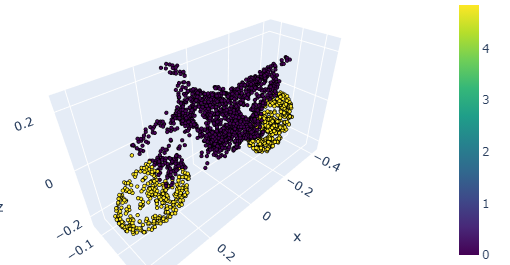}
        \subcaption{\label{fig:seg_cluster_usecase_example_b} Saliency map of (a)}
    \end{subfigure}
    \vfill
    \begin{subfigure}[b]{0.48\linewidth}
        \centering
        \includegraphics[width=\linewidth]{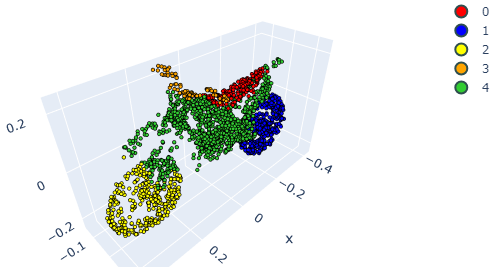}
        \subcaption{\label{fig:seg_cluster_usecase_example_c}Segmentation+Clustering}
    \end{subfigure}
    \hfill
    \begin{subfigure}[b]{0.48\linewidth}
        \centering
        \includegraphics[width=\linewidth]{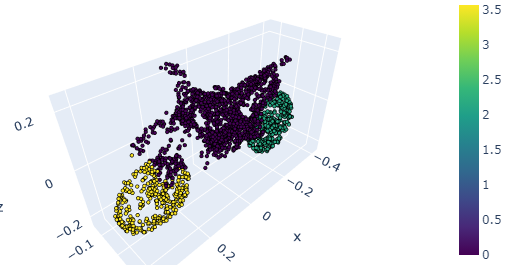}
        \subcaption{\label{fig:seg_cluster_usecase_example_d} Saliency map of (c)}
    \end{subfigure}
    \caption{\label{fig:seg_cluster_usecase_example} Saliency maps of point clouds representing motorbikes using the \emph{absence of feature} mechanism.}
\end{figure}
\autoref{fig:seg_cluster_usecase_example} shows an example of saliency maps generated for point cloud data representing a motorbike. The division of input data into multiple clusters is performed using both segmentation and segmentation+clustering mechanisms. We observe that the wheels are clustered as one segment in \autoref{fig:seg_cluster_usecase_example_a} while the segmentation+clustering mechanism manages to cluster them as separate segments as shown in \autoref{fig:seg_cluster_usecase_example_c}. We observe that the saliency map generated using the segmentation mechanism indicates high importance for wheels. However, it does not provide any information regarding which wheel is more influential. This is addressed by the segmentation+clustering mechanism which enables us to introduce more specific perturbations into the input data using individual wheels. The saliency map generated using this mechanism is shown in \autoref{fig:seg_cluster_usecase_example_d}. It shows the front wheel to have more influence on the output class score compared to the rear wheel. This is useful mainly because the wheels of the motorbike are not identical and their locations with respect to the remaining features in the input point clouds are also different. Therefore, these wheels are expected to have different levels of influence on the output class score which is highlighted in \autoref{fig:seg_cluster_usecase_example_d}.  

\subsection{Performance analysis}
We analyze the performance of our methods using the ground truth of the segmentation task and noisy point cloud data as the input for the classification model in the pipeline.

\subsubsection{Ground truth}
\begin{figure}[btp]
    \centering
    \begin{subfigure}[b]{0.46\linewidth}
        \centering
        \includegraphics[width=\linewidth]{figures/sal_maps/Airplane/Airplane_13.png}
        \subcaption{\label{fig:Airplane_13_XAI}Input}
    \end{subfigure}
    \hfill
    \begin{subfigure}[b]{0.46\linewidth}
        \centering
        \includegraphics[width=\linewidth]{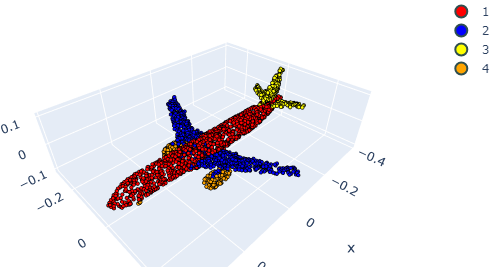}
        \subcaption{\label{fig:Airplane_13_GT_XAI}Segmentation GT}
    \end{subfigure}
    \vfill
    \begin{subfigure}[b]{0.46\linewidth}
        \centering
        \includegraphics[width=\linewidth]{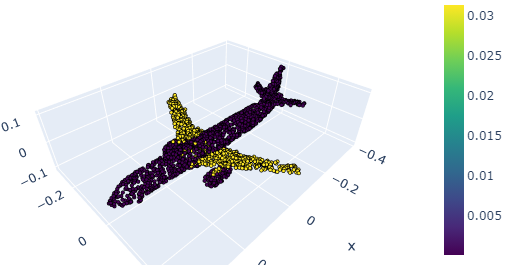}
        \subcaption{\label{fig:Airplane_13_GT_sal_map_XAI}Saliency map with GT}
    \end{subfigure}
    \hfill
    \begin{subfigure}[b]{0.46\linewidth}
        \centering
        \includegraphics[width=\linewidth]{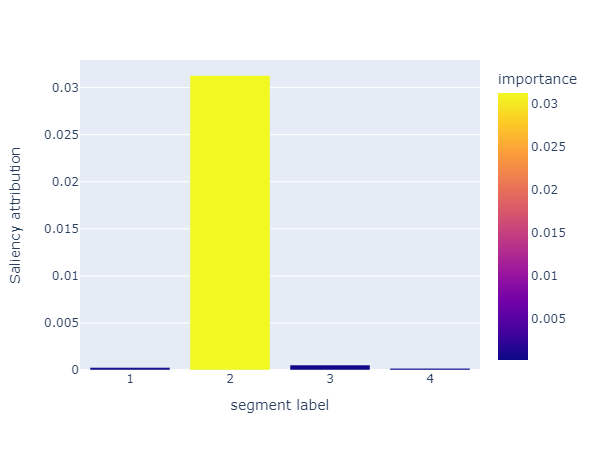}
        \subcaption{\label{fig:Airplane_13_GT_sal_map_barplot_XAI}Saliency attributions}
    \end{subfigure}
    \vfill
    \begin{subfigure}[b]{0.46\linewidth}
        \centering
        \includegraphics[width=\linewidth]{figures/sal_maps/Airplane/Airplane_13_sal_map.png}
        \subcaption{\label{fig:Airplane_13_sal_map_XAI}Saliency map without GT}
    \end{subfigure}
    \hfill
    \begin{subfigure}[b]{0.46\linewidth}
        \centering
        \includegraphics[width=\linewidth]{figures/sal_maps/Airplane/Airplane_13_sal_map_barplot.png}
        \subcaption{\label{fig:Airplane_13_sal_map_barplot_XAI}Saliency attributions}
    \end{subfigure}
    \caption{\label{fig:XAI_ground_truth}Saliency maps produced with ((c) \& (d)) and without ((e) \& (f)) using the segmentation ground truth labels using the \emph{absence of feature} mechanism. Note: Refer to the segmentation ground truth figure (b) for corresponding parts represented along the x-axis in the bar plot (d) \& (f).}
\end{figure}

This analysis corresponds to the saliency maps generated by our method based on the segments present in the ground truth (GT) segmentation. The ground truth is used in the third stage of our XAI pipeline which is used to perturb the input point cloud data. \autoref{fig:XAI_ground_truth} shows an example of the saliency map generated for the ground truth of an input instance representing an airplane. We use this scenario because the ground truth is the "perfect output" of the segmentation model. In other words, ground truth would be the output of the segmentation model if it had 100\% accuracy. Therefore, it is important to analyze the performance of segmentation models in generating the saliency maps as they are a central part of our proposed method. 

An example of the saliency maps produced using ground truth and the output of the segmentation model is shown in \autoref{fig:XAI_ground_truth}. We compare the saliency maps to analyze how the inaccuracy of a segmentation model affects the saliency attributions of the segments in the input data. We observe that the segmentation model manages to produce saliency maps ((e) \& (f)) similar to those produced using the segmentation ground truth data ((c) \& (d)). This highlights the ability of segmentation models to generate meaningful segments with high accuracy, which leads to the production of meaningful saliency maps. 

\subsubsection{Noisy input}
During the training process, a classification model learns to produce a desired output by tuning its parameters based on the input instance provided and its corresponding ground truth. At the end of the training process, the model parameters are tuned well enough to produce the desired output for a subset of the training dataset (assuming the model does not reach 100\% accuracy). However, to analyze the model more effectively, it is important to test its performance on input instances that the model has not seen during its training process. 

\begin{figure}[t]
    \centering
    \begin{subfigure}[b]{0.46\linewidth}
        \centering
        \includegraphics[width=\linewidth]{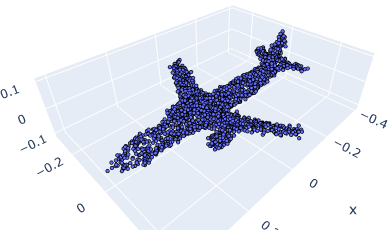}
        \subcaption{\label{fig:Airplane_13_noisy}Noisy input}
    \end{subfigure}
    \hfill
    \begin{subfigure}[b]{0.46\linewidth}
        \centering
        \includegraphics[width=\linewidth]{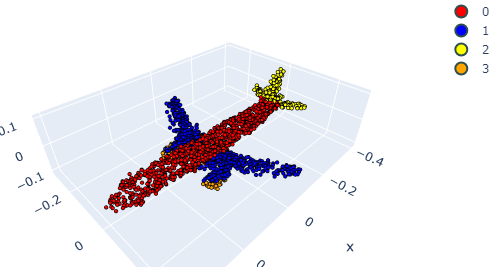}
        \subcaption{\label{fig:Airplane_13_noisy_seg}Segmentation output}
    \end{subfigure}
    \vfill
    \begin{subfigure}[b]{0.46\linewidth}
        \centering
        \includegraphics[width=\linewidth]{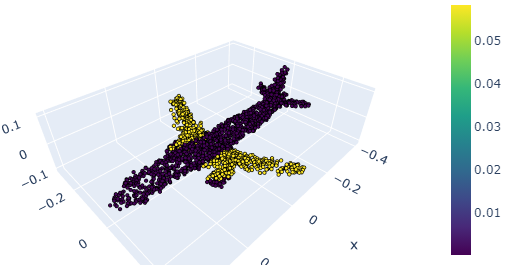}
        \subcaption{\label{fig:Airplane_13_noisy_sal_map}Saliency map}
    \end{subfigure}
    \hfill
    \begin{subfigure}[b]{0.46\linewidth}
        \centering
        \includegraphics[width=\linewidth]{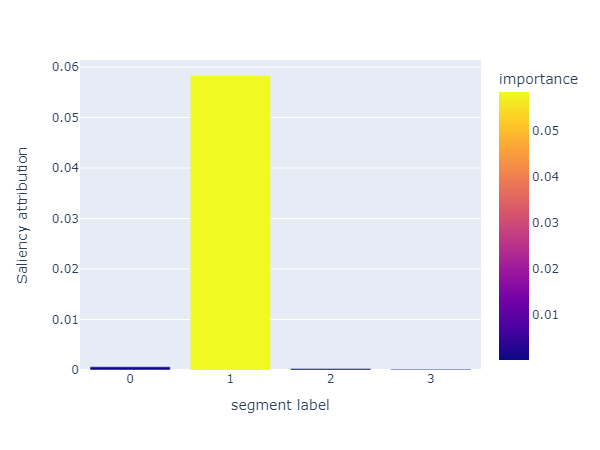}
        \subcaption{\label{fig:Airplane_13_noisy_sal_map_barplot}Saliency attributions of segments.}
    \end{subfigure}
    \caption{\label{fig:XAI_noisy_input}Noisy input with 5\% noise and its corresponding saliency attributions produced by our proposed method.Note: Refer to the 'segmentation output' figure (b) for corresponding parts represented along the x-axis in the bar plot (d).}
\end{figure}



One of the most common mechanisms of generating new examples for testing AI models is by adding noise to the available data instances. We use this mechanism to test the classification model as well as our XAI mechanism. We introduce noise into the input data instances and analyze the saliency maps generated. An example of this analysis is shown in \autoref{fig:XAI_noisy_input}. We observe that our method produces similar saliency attributions for a noisy input data instance that is generated by adding 5\% noise to the actual input data (see \autoref{fig:Airplane_13_sal_map_XAI} and \autoref{fig:Airplane_13_sal_map_barplot_XAI}). We observed that the method produces saliency maps with minute variations up to 10\% noise level. However, higher levels of noise magnitude lead to bigger changes in the input data and, therefore, lead to the generation of incorrect segmentation, leading to incorrect saliency maps. This indicates that the classification model is robust to noise in the input data and also highlights the performance of segmentation models that are a major part of our proposed method.

\subsection{Limitations}\label{limitations}
\begin{figure}[bt]
    \centering
    \includegraphics[width=\linewidth]{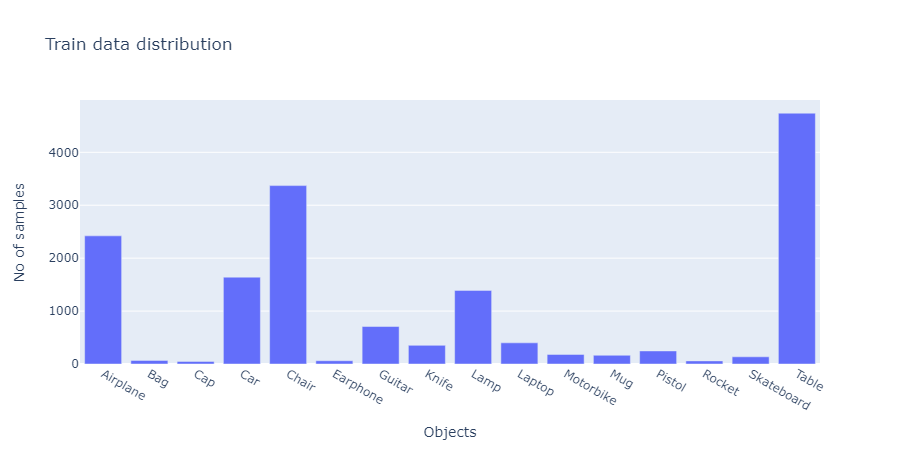}
    \caption{Distribution of the training data.}
    \label{fig:train_data_details}
\end{figure}

One of the limitations of our proposed method is associated with the use of a segmentation algorithm. This limitation is the possible inaccuracy of the segmentation model that is trained on the point cloud data. The segmentation models used in this work have accuracy values in the range of 65\%-80\%. This is mainly due to two reasons. The first reason corresponds to the imbalance in the dataset. The dataset contains varying numbers of samples representing individual 3D model types with some of them having less than 100 samples as shown in \autoref{fig:train_data_details}. This makes learning difficult for the segmentation models.
The second reason corresponds to the structural information associated with these 3D models. Some point clouds represent simple 3D models such as chairs, tables, and laptops which make it easier for the corresponding segmentation models to learn the segmentation task. However, point clouds representing complex 3D models such as motorbikes, cars and airplanes make learning more difficult for the AI models. 

The second limitation is also associated with the dataset. This limitation is the requirement of a labeled dataset for segmentation in case we decide to add another 3D object or category to the classification task. This is due to the use of segmentation models that are trained for the segmentation of point clouds of specific categories.

The last limitation corresponds to the dependency of the XAI method on the output label of the classification model when working without human input in the pipeline. Currently, the method uses the classification output to find the corresponding segmentation model. However, when a classification model incorrectly classifies the input data, it will lead to selecting a segmentation model that is inappropriate for the input data. However, human-in-the-loop can easily resolve this problem with the user selecting the segmentation model based on the input point cloud data. 

\section{Conclusion}\label{conclusion}
In this paper, we proposed a segmentation-based XAI method for understanding the decision-making process of classification models working on point cloud data. The proposed method is based on a perturbation mechanism. It specifically uses meaningful segments to introduce perturbations and thus, produces more meaningful saliency maps. We used two types of perturbation mechanisms to generate explanations with two different perspectives. This allows users to gain better insight into the decision-making process of a classification model and the information carried by each segment in the input data. For the segmentation task, we proposed two mechanisms that leverage segmentation models and clustering algorithms to generate saliency maps. We also proposed a new point-shifting mechanism for the perturbation, to improve explainability. Applying the method to several representative examples, we highlighted the usefulness of our proposed method and analyzed its performance using different input data instances.
The proposed method is model-agnostic and therefore can be used to explain any classification model working on point cloud data, irrespective of the model architecture. 

Our future work will be to address the limitations mentioned in \autoref{limitations}.

\bibliographystyle{plainurl}
\bibliography{literature}

\end{document}